\documentclass[lettersize,journal]{IEEEtran}
\usepackage{cite}
\usepackage{amsmath,amsfonts}
\usepackage{array}
\usepackage{amssymb}% http://ctan.org/pkg/amssymb
\usepackage{pifont}% http://ctan.org/pkg/pifont
\usepackage{textcomp}
\usepackage{stfloats}
\usepackage{url}
\usepackage{verbatim}
\usepackage{graphicx}
\usepackage{algorithm}
\usepackage{algpseudocode}
\usepackage{caption}
\usepackage{color}
\usepackage[flushleft]{threeparttable}
\usepackage{subcaption}
\def\BibTeX{{\rm B\kern-.05em{\sc i\kern-.025em b}\kern-.08em
    T\kern-.1667em\lower.7ex\hbox{E}\kern-.125emX}}
\usepackage{balance}
\usepackage{hyperref}
\usepackage{adjustbox}
\usepackage{mathtools}

\usepackage{tikz}
\usetikzlibrary{shapes.geometric, arrows}
\newcommand{\fedsyn}{\texttt{TrajSyn}}
\newcommand{\fedsynfed}{\texttt{TrajSynFed}}

\begin{document}

% \title{FedSyn: Privacy-Preserving Synthetic Data Generation for Adversarially Robust Federated Learning}
\title{TrajSyn: Privacy-Preserving Dataset Distillation from Federated Model Trajectories for Server-Side Adversarial Training}

\author{Mukur Gupta, Niharika Gupta, Saifur Rahman, \IEEEmembership{Member, IEEE}, \\ Shantanu Pal \IEEEmembership{Senior Member, IEEE}, and Chandan Karmakar, \IEEEmembership{Member, IEEE}
% \thanks{Manuscript received}

\thanks{M. Gupta, was with the Computer Science Department, Columbia University, New York, USA (e-mail: mukur.gupta@columbia.edu)}
\thanks{N. Gupta, is with the School of Computer Science and Engineering, Vellore Institute of Technology, Vellore, India (e-mail: 
niharikagupta1203@gmail.com)}
\thanks{S. Rahman, S. Pal, and C. Karmakar are with the School of Information Technology, Deakin University, Melbourne, VIC 3125, Australia (e-mail: 
saifur.rahman@deakin.edu.au, shantanu.pal@deakin.edu.au, karmakar@deakin.edu.au)}
}

% \markboth{IEEE TRANSACTIONS ON TRANSACTIONS, December~2024}%
% {How to Use the IEEEtran \LaTeX \ Templates}

\maketitle

\begin{abstract}
Deep learning models deployed on edge devices are increasingly used in safety-critical applications. However, their vulnerability to adversarial perturbations poses significant risks, especially in Federated Learning (FL) settings where identical models are distributed across thousands of clients. While adversarial training is a strong defense, it is difficult to apply in FL due to strict client-data privacy constraints and the limited compute available on edge devices. In this work, we introduce \fedsyn, a privacy-preserving framework that enables effective server-side adversarial training by synthesizing a proxy dataset from the trajectories of client model updates, without accessing raw client data. We show that \fedsyn{ }consistently improves adversarial robustness on image classification benchmarks with no extra compute burden the client device.
\end{abstract}


\section{Introduction}
Recent advances in the Internet of Things (IoT), edge computing, and deep learning have enabled the deployment of sophisticated models directly on resource-constrained edge devices\cite{10502625}. Deep learning systems now operate on smartphones, home assistants, autonomous drones, and vehicular platforms, processing local sensory data in real time. As the number of edge-enabled devices continues to grow, we have seen a rapid growth in on-device workloads, reflecting an industry push toward decentralized, privacy-preserving computation.
To preserve user privacy and reduce communication overhead, edge deployments avoid sending raw sensory data to the cloud. Instead, models are trained or fine-tuned locally, and Federated Learning (FL) \cite{mcmahan2023communicationefficient} is most commonly used for training such models on user data that cannot leave the device. This training mechanism is frequently used in safety-critical applications, e.g.,autonomous driving, smart-city infrastructure, and healthcare monitoring, where robust and reliable predictions are essential.

However, extensive prior work has demonstrated that deep learning models are highly vulnerable to subtle input perturbations that can cause models to mispredict with high confidence \cite{szegedy2014intriguing, madry2019deep, carlini2017adversarial, wong2018polytope, li2019noise, zhang2018efficient, cohen2019certified, papernot2016distillation, zou2023universal}. These vulnerabilities have been observed across a wide range of domains. For example, small physical or digital modifications can mislead traffic-sign recognition systems \cite{pavlitska2023adversarialattackstrafficsign}, and carefully crafted perturbations can inject harmful behaviors into coding agents \cite{storek2025xoxostealthycrossorigincontext}. These threats become even more severe in a white-box setting, where an adversary has full access to model parameters and can craft highly effective gradient-based or even universal attacks.
In FL settings, the attack surface for white-box adversarial attacks becomes especially large because vendors often deploy identical model architectures across thousands of client devices. As a result, a single malicious client with full access to the shared global model can craft effective adversarial examples that typically transfer to all other devices running the same model. Thus, any attack vector discovered by one compromised participant can be leveraged at scale across the entire deployment.

To defend against such adversarial attacks, prior research (e.g., \cite{eloundou2023gpts, madry2019deep, gupta2025advsummadversarialtrainingbias, wong2018polytope}) has shown the effectiveness of adversarial training, wherein models are trained on adversarially perturbed examples to increase their robustness to input perturbations. Explicitly exposing the model to worst-case perturbations during training, adversarial training significantly improves robustness against both gradient-based and black-box attacks.

However, applying adversarial training in an FL setting introduces two key challenges. First, privacy constraints in FL prohibit the server from accessing raw client data, which is required to generate adversarial examples, thus preventing the server from performing adversarial training directly. Second, computational limitations on edge devices make adversarial training prohibitively costly for clients \cite{liang2024acfl}. Generating adversarial examples typically requires multiple forward and backward passes per input, which leads to substantial additional computation and energy consumption. Many real-world client devices, e.g., smartphones, IoT sensors, and embedded automotive systems, lack the resources to support this workload, rendering client-side adversarial training impractical. Together, these constraints create a significant gap: adversarial training is necessary for robustness, yet neither the server nor the clients can feasibly perform it within the traditional FL paradigm.

\begin{figure*}[!htp]
    \centering
    \includegraphics[scale=0.75]{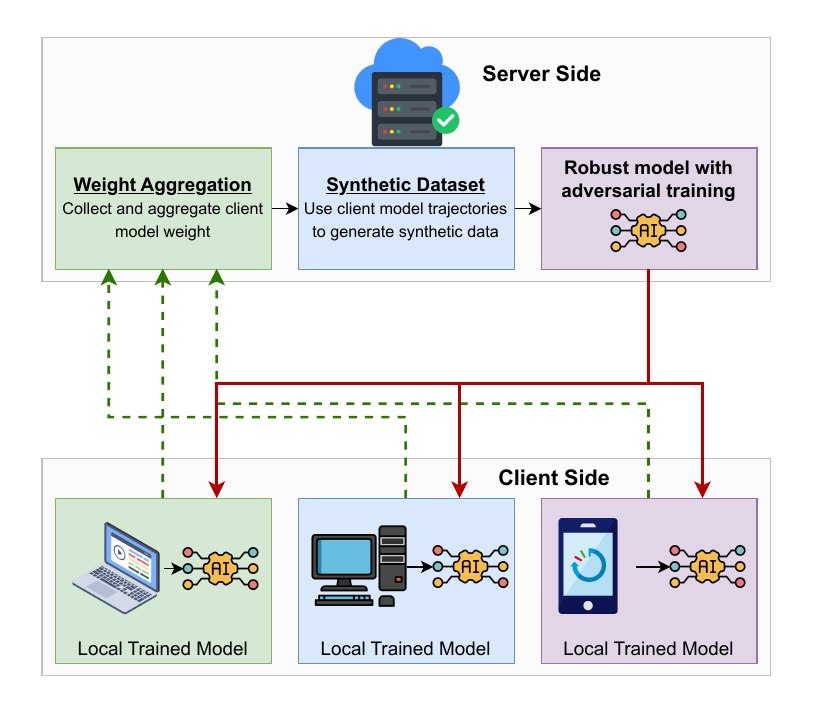}
    \caption{Overview of the workflow in this \fedsyn{} setup. Each device trains a local model on its own private data. The server gathers the client model weights, shown by the green dotted lines, and combines them. It then uses the model trajectories to build a synthetic dataset. This dataset is used for adversarial training on the server to produce a more robust global model. The improved model is then returned to the clients, shown by the red solid lines.}
    \label{fig:feddist}
\end{figure*}

In this paper, we propose \fedsyn~(Model \underline{Traj}ectory \underline{Syn}thesis), a novel method that enables server-side adversarial training in FL while fully preserving client data privacy. As illustrated in Fig.~\ref{fig:feddist}, \fedsyn~begins with a standard FL training in which each client trains locally and sends its model updates to the server. Instead of discarding these updates after aggregation, the server logs the sequence of received model weights across communication rounds, forming a set of client-specific model training trajectories. Using these trajectories, \fedsyn~synthesizes a server-side dataset that approximates the distribution of client data, without ever accessing or reconstructing any raw client inputs. This synthetic dataset provides the server with a surrogate on which it can perform compute-intensive adversarial training. After adversarially fine-tuning the global model on the synthesized data, the server redistributes the robust model back to all clients. In doing so, \fedsyn~transfers the heavy computational burden of adversarial training entirely to the server while maintaining the privacy guarantees in FL.
We demonstrate the effectiveness of \fedsyn~on an image classification benchmark. Our contributions are twofold:
\begin{itemize}
    \item A new dataset-distillation framework for FL that generates a high-quality, privacy-aware synthetic dataset suitable for applications beyond adversarial robustness.
    \item Empirical evidence that \fedsyn~substantially improves adversarial robustness over standard FL training, while imposing no additional computational load on clients.
\end{itemize}

The rest of the paper is organized as follows. Section~\ref{related-work} reviews the relevant literature. Section~\ref{sec: setting} introduces the problem setting, and Section~\ref{sec: approach} describes our proposed method. Experimental design and results are presented in Sections~\ref{sec: experiments}. Section \ref{results-discussions} provides the experimental results and discussions. Finally, Section \ref{conclusion-future-work} concludes the paper with future work.

\section{Related Work}
\label{related-work}

\subsection{Adversarial Training}
The problem of adversarial attacks has been widely studied in deep learning, where small changes in model input - which are indistinguishable to the human eye - can cause a model to completely flip its output with high confidence \cite{szegedy2014intriguing}.
Several lines of defences to adversarial attacks have been explored over the years \cite{carlini2017adversarial} including network verification with polytope approximation \cite{wong2018polytope, zhang2018efficient}, randomized smoothing \cite{li2019noise, cohen2019certified}, lipshitz function regularization \cite{anil2019sorting, hein2017formal, peck2017lower}, model distillation \cite{papernot2016distillation} and adversarial training \cite{madry2019deep}. While approaches like polytope approximation and Lipschitz function regularisation provide robustness guarantees, they are either applicable to simple ReLu networks \cite{croce2019provable} or fail to scale up as model sizes grow larger \cite{pmlr-v162-bab}. While providing no theoretical bounds, adversarial training provides empirical robustness, which in practice is good enough for most of the practical non-critical applications. With the advent of Large Language Models (LLMs), attacks have evolved beyond the classification models to attacking the LLMs in blackbox and whitebox settings by perturbing the input prompts \cite{zou2023universal, maus2023blackbox, storek2025xoxostealthycrossorigincontext, cui2024recentadvancesattackdefense}.

Several prior works have explored adversarial robustness in the FL setting. For example, FAT \cite{zizzo2020fat} performs adversarial training locally on each client during every communication round before sending model updates to the server. While this approach improves the robustness of the aggregated global model, it imposes substantial computational overhead on client devices, an impractical requirement for many real-world edge deployments with limited resources. In contrast, \fedsyn~shifts the burden of adversarial training to the server, which typically has significantly greater computational capacity. In our evaluations, we compare the performance of \fedsyn~against FAT.

\subsection{Synthetic Data Generation in FL}
Several approaches to data generation have been explored by the prior research, e.g., data distillation by learning soft labels \cite{sucholutsky2021soft}, introducing differentiable data augmentation \cite{zhao2021dataset}, and gradient matching between weights of a network trained on a synthetic dataset and with network trained on the original dataset \cite{zhao2020dataset}. In the context of FL, \cite{zhou2020distilled} sends the statistical distribution of each feature in the data to the server along with the client model. However, this approach is limited to tabular data and doesn't transfer well to non-structured image or text datasets. 

Proposal \cite{song2023federated} performs data distillation on each client machine, then send this to the server, where all the datasets are aggregated, and a server model is trained on the aggregated distilled dataset. Proposal \cite{zhou2020distilled} uses a similar setup for one-shot learning on the server. While distillation on the client machine gives some performance boost on the server model, but requires the client to perform the computationally heavy data distillation step. Furthermore, sharing the distilled dataset, which is initialised on the client data, violates privacy constraints in some scenarios. Our synthetic data generation strategy differs from these approaches in two ways. Firstly, the data generation process incurs no additional computation on the Client side, and we use model weights instead of client data for generating the dataset.

\textit{In summar}y, \fedsyn~differs from prior work by eliminating all client-side overhead and avoiding any sharing of real or distilled data. Instead of requiring clients to run costly adversarial training or data distillation, \fedsyn~uses only the model weight trajectories already exchanged during FL to generate synthetic data on the server. This enables server-side adversarial training while fully preserving FL privacy constraints and keeping client computation unchanged.

\section{Problem Setting}
\label{sec: setting}

FL \cite{mcmahan2023communicationefficient} is a privacy-preserving paradigm in which a shared global model is trained collaboratively across multiple client devices, while ensuring that each client’s raw data remains local. Consider $N$ clients $\{C_1, C_2\dots C_N\}$ each holding a private dataset $\mathcal{D}_{C_1}, \dots, \mathcal{D}_{C_N}$ that never leaves the device. Each client $C_i$ trains a local model $M_{C_i}$ on its own dataset $\mathcal{D}_{C_i}$. Since individual datasets may be too small or heterogeneous to support strong generalization, clients periodically send their model parameters, not the data, to a central server. The server aggregates these parameters \cite{qi2023model} to produce an updated global model, which is then redistributed to all clients for the next round of local training. This process repeats for $T$ communication rounds, enabling clients to benefit from collective knowledge while preserving strict data locality and privacy.

\section{Proposed Approach}
\label{sec: approach}
Based on the problem setting described in Section~\ref{sec: setting}, in Section~\ref{sec: threat model}, we describe the threat model, followed by emphasizing the dual challenges posed by strict privacy constraints and the limited computational resources typical of edge devices. Section~\ref{sec: fedsyn} then introduces our proposed approach, \fedsyn, which constructs a server-side synthetic dataset that approximates the distribution of client data without accessing any raw local information. This enables a final round of adversarial training performed entirely on the server, yielding a more robust global model while fully preserving the privacy guarantees of FL. The complete \fedsyn~algorithm is summarized in Algorithm~\ref{alg:alg}.

\subsection{Threat Model} 
\label{sec: threat model}

In FL, the global model shared across clients remains vulnerable to adversarial attacks \cite{szegedy2014intriguing, kumar2023impact}, due to the white-box accessibility provided to every participating device. After each communication round, all clients receive the same updated global model, meaning that a single compromised or malicious client can obtain full knowledge of the model architecture and parameters. Under this white-box threat model, any adversarial example crafted on one device can transfer well to all others running the same model. As large vendors often deploy identical FL-trained models across thousands or even millions of edge devices, this uniformity dramatically expands the attack surface, enabling adversaries to exploit vulnerabilities at scale.

\subsection{Adversarial Training} 
Adversarial training \cite{madry2019deep} is the most widely used defense for improving model robustness against adversarial attacks. The key idea is to augment the training process with adversarial examples that are inputs intentionally perturbed to cause misclassification at inference time. These adversarial examples are generated by attacking the current model state and identifying inputs that are most likely to cause the model to flip the true label.

We adopt the standard min–max optimization objective from adversarial training \cite{madry2019deep}, which uses Projected Gradient Descent (PGD) as the strongest first-order adversary. Empirically, training the model on PGD-generated adversarial samples is sufficient to achieve robustness. The perturbations used to create adversarial examples are typically constrained by an \( \ell_p \)-norm bound to ensure that the semantic content of the original input is preserved. 

Formally, an adversarial example \(x^{\text{adv}}\) for a given input sample \(x\) is generated as:
\begin{align}
\label{eq:adversarial step}
    x^{\text{adv}} = x + \eta \frac{\partial \mathcal{L}}{\partial x}, 
    \qquad \|x^{\text{adv}} - x\|_{p} \leq \epsilon,
\end{align}
where, \(\mathcal{L}\) denotes the model loss function, computing the gradient requires white-box access to model parameters, \(\eta\) is the step size controlling the perturbation magnitude, and \(\epsilon\) specifies the allowable \( \ell_p \)-norm bound on the adversarial perturbation.

In the adversarial training process, Equation~\ref{eq:adversarial step} is applied repeatedly for $\lambda$ steps to generate $x^{adv}$. After $\lambda$ steps, the final $x^{adv}$ is then used after each epoch to complete a round of adversarial training.
Each generation of $x^{adv}$ using Equation~\ref{eq:adversarial step} involves a forward model pass followed by gradient computation (with respect to the input $x$), which induces a computational overhead equivalent to one epoch. This training process, therefore, adds an additional overhead of $\lambda \times \text{num\_epochs}$ to the overall training process. This overhead becomes increasingly difficult for a client device. The server, on the other hand, may be a GPU cluster capable of running the compute-heavy adversarial training process. However, FL prevents the server from accessing any client data $x$, which would be required for an adversarial round.

In adversarial training, Equation~\ref{eq:adversarial step} is applied iteratively for \(\lambda\) steps to generate the final adversarial example \(x^{\text{adv}}\). After these \(\lambda\) iterations, the resulting \(x^{\text{adv}}\) is used at the end of each epoch to perform a round of adversarial updates. Each iteration of Equation~\ref{eq:adversarial step} requires both a forward pass of the model and a backward pass to compute gradients with respect to the input \(x\). Consequently, generating a single adversarial example incurs computational cost \(\lambda\) times that of training for an epoch. Over the full training process, adversarial training introduces an overhead of approximately \(\lambda \times \text{num\_epochs}\), which rapidly becomes prohibitive for resource-constrained client devices in FL.

Although the central server in an FL system typically has substantial computational resources, often a GPU-equipped cluster capable of performing compute-intensive adversarial training, FL’s strict privacy constraints prevent the server from accessing any client inputs \(x\), which are precisely the quantities needed to generate adversarial examples. This creates a problem where clients possess the data required for adversarial example generation but lack the computational capacity to produce them, while the server has the computational capability but is not permitted to access client data.
Next, we describe how \fedsyn\ leverages the client model parameters received after each communication round to construct a synthetic dataset \(\mathcal{D}_{\text{syn}}\) that approximates the clients' data distribution without accessing any raw client data. This synthesized dataset \(\mathcal{D}_{\text{syn}}\) is then used to complete a round of adversarial training on the server.

\begin{table}[t]
\caption{Clean accuracy in FL setting with variation in image initialization. Experiments on the CIFAR-10 dataset with the ConvNet Model. The complete CIFAR-10 dataset size is 50,000. *Accuracy taken from \cite{cazenavette2022dataset}}
  \label{tab:img init}
  \centering
\begin{adjustbox}{width=0.9\columnwidth}  
  \begin{tabular}{lcccc}
    \hline
     Expert &\# Distilled Imgs & \# Experts & Img Initialization & Accuracy\\
    \hline
    Federated & 500 & 11 & Real & 58\\
    Federated & 500 & 11 & Gaussian & 28\\
    \hline
    Centralized & 500 & 100 & Real & 71*\\
    %FL & 7500 & 11 & Gaussian & 49\\
    \hline
  \end{tabular}
\end{adjustbox}  
%\vskip -0.1in
\end{table}

\subsection{Data Distillation}
\label{sec:apporach data distillation}
Dataset distillation aims to construct a small synthetic dataset that enables a model to achieve performance comparable to training on the full real dataset. An approach by \cite{cazenavette2022dataset} accomplishes this by exploiting \emph{training trajectories} of a model as expert supervision for a student network.
In this method, a teacher model $\mathcal{M}$ is trained on the real dataset $\mathcal{D}_{\text{real}}$ from multiple random initializations. During training, the model parameters at each epoch are recorded. For a single initialization, the sequence of parameters
\(
\{\theta_1, \theta_2, \dots, \theta_T\}
\)
over $T$ epochs forms a training trajectory $\tau_i$. Collectively, the set $\{\tau_i\}$ serves as a collection of expert trajectories.

A student network $\mathcal{S}$ is then trained using these expert trajectories. The synthetic distilled dataset $\mathcal{D}_{\text{dis}}$ is initialized either from a small subset of $\mathcal{D}_{\text{real}}$ or from Gaussian noise. To begin a distillation step, a trajectory $\tau_i$ is sampled, and the student model is initialized with the expert parameters at a randomly chosen epoch $\theta^i_t$. The student is trained for $N$ epochs on $\mathcal{D}_{\text{syn}}$, minimizing the task-specific loss (e.g., cross-entropy for classification). Let $\hat{\theta}_N$ denote the student parameters after these $N$ epochs.

To update the synthetic data, the method minimizes the discrepancy between the student parameters $\hat{\theta}_N$ and the expert parameters $\theta^i_{t+N}$, using the following normalized trajectory-matching loss:
\begin{align}
\label{eq:dist loss}
\mathcal{L}
= 
\frac{\left\| \hat{\theta}_N - \theta^i_{t+N} \right\|_2^2}
     {\left\| \hat{\theta}_0 - \theta^i_{t+N} \right\|_2^2}.
\end{align}
Optimizing this loss updates $\mathcal{D}_{\text{syn}}$ such that training the student on $\mathcal{D}_{\text{syn}}$ mimics the parameter evolution that would be obtained by training on $\mathcal{D}_{\text{real}}$. Repeating this process for $M$ distillation steps yields a synthetic dataset $\mathcal{D}_{\text{syn}}$ that captures the learning dynamics of the real data.

Although both expert and student networks share the same architecture during distillation, \cite{cazenavette2022dataset} shows that the resulting synthetic dataset is largely model-agnostic and transfers well across different architectures, demonstrating the broad application of trajectory-based dataset distillation.

\subsection{\fedsyn: Federated Synthesis}
\label{sec: fedsyn}

Building on the data distillation strategy described in Section~\ref{sec:apporach data distillation}, we propose \fedsyn, a framework that adapts trajectory-based dataset distillation to the FL setting. During FL training, each client \(C_i\) trains a local model \(\mathcal{M}_{C_i}\) and sends its model parameters \(\theta^{C_i}_t\) to the server at every communication round $t$ for aggregation. The sequence of parameters received from each client naturally forms a trajectory \(\tau_i\), and the collection of all such trajectories \(\{\tau_i\}\) serves as the expert set for distillation. After the completion of FL training, these client trajectories are used by the server to synthesize a dataset \(\mathcal{D}_{\text{syn}}\) that approximates the underlying distribution of the combined client data. This synthetic dataset is then used to perform multiple rounds of adversarial training on the server, enabling robustness improvements without accessing any raw client data. We note that \(\mathcal{D}_{\text{syn}}\) produced by \fedsyn\ may support additional downstream applications beyond adversarial training, which we leave for future work.

A key distinction between \fedsyn and the standard dataset distillation procedure in Section~\ref{sec:apporach data distillation} lies in the nature of the training trajectories. In typical data distillation, all expert trajectories are obtained by training on the full real dataset \(\mathcal{D}_{\text{real}}\). In contrast, the trajectories \(\theta^{C_i}_t\) in \fedsyn\ are generated from client datasets that are smaller and drawn from different iid. subsets of \(\mathcal{D}_{\text{real}}\). As a result, initializing \(\mathcal{D}_{\text{syn}}\) with Gaussian noise, as shown in \cite{cazenavette2022dataset}, yields poor distillation performance due to the heterogeneous nature of these trajectories (see Table~\ref{tab:img init}).

To address this challenge, \fedsyn\ assumes the server participates in FL as an additional client, possessing or generating a small local dataset sampled from the same underlying distribution. Under the standard iid. assumption of FL, this allows \(\mathcal{D}_{\text{syn}}\) to be initialized using a small subset of server data rather than random noise, improving stability and fidelity during distillation. We present the full \fedsyn~procedure in Algorithm~\ref{alg:alg}.

\begin{algorithm}[t]
    \caption{Dataset Distillation in FL}
    \label{alg:alg}
    \begin{algorithmic}[1]
        \Require $\{C_1, C_2, \dots, C_N\}$: participants in Federated Learning
        \Require $\{M_{C_1}, M_{C_2}, \dots, M_{C_N}\}$: client models
        \Require $\lambda, \eta, \epsilon$: adversarial training parameters
        \Require $T$: epochs for FL; $E$: epochs to train client model; $E^{adv}$: epochs for adversarial training
        \Require $\mathcal{D}_{C_1}, \dots, \mathcal{D}_{C_N}$: local client datasets, such that $\bigcup_i \mathcal{D}_{C_i} = \mathcal{D}_{\mathrm{real}}$

        \State Initialize distilled data $\mathcal{D}_{\mathrm{dis}}$
        \State Initialize empty expert trajectories for each $C_i$: $\{\tau_i = \{\}\}$

        \For{$t = 1$ to $T$}
            \State $\triangleright$ Train client models $\{M_{C_i}\}$ on $\mathcal{D}_{C_i}$
            \State $\triangleright$ Send model parameters $\theta_t^{C_i}$ of each $C_i$ to server
            \State $\triangleright$ For all $C_i$, append $\theta_t^{C_i}$ to $\tau_i$
            \State $\triangleright$ Aggregate parameters $\sum_i \theta_t^{C_i} / N$ and update each $M_{C_i}$
        \EndFor

        \For{each distillation step}
            \State $\triangleright$ Sample expert trajectory: $\tau^* \sim \{\tau_i\}$ with $\tau_i = \{\theta_t^{C_i}\}_{t=0}^T$
            \State $\triangleright$ Choose random start epoch $t \leq T$
            \State $\triangleright$ Initialize student network $\mathcal{S}$ with expert parameters:
            \State \hspace{1.5em} $\hat{\theta}_0 \coloneqq \theta_t^{C_i}$
            \For{$n = 0$ to $E - 1$}
                \State $\triangleright$ Update $\mathcal{S}$ w.r.t. classification loss:
                \State \hspace{1.5em} $\hat{\theta}_{n+1} = \hat{\theta}_{t} - \alpha \nabla \ell(\mathcal{D}_{\mathrm{dis}}; \hat{\theta}_{t+n})$
            \EndFor
            \State $\triangleright$ Update $\mathcal{D}_{\mathrm{dis}}$ with respect to $\mathcal{L}$:
            \State \hspace{1.5em} $\mathcal{L} = \dfrac{\|\hat{\theta}_{N} - \theta_{t+N}^{C_i}\|_2^2}{\|\theta_{t}^{C_i} - \theta_{t+N}^{C_i}\|_2^2}$
        \EndFor

        \State Final server model $\mathcal{M}_S = \sum_i \theta_T^{C_i} / N$

        \For{$n = 0$ to $E^{adv} - 1$}
            \State $\triangleright$ Adversarially train server model
            \State \hspace{1.5em} $\mathcal{M}_S \coloneqq \text{adv\_train}(\mathcal{M}_S, \lambda, \eta, \epsilon)$
        \EndFor

        \Ensure Distilled data $\mathcal{D}_{\mathrm{dis}}$, adversarially robust model $\mathcal{M}_S$
    \end{algorithmic}
\end{algorithm}

\section{Experiments}
\label{sec: experiments}
\subsection{Datasets}
We evaluate \fedsyn~on an image classification task using the CIFAR-10 dataset \cite{krizhevsky2009learning}, which consists of 50{,}000 training and 10{,}000 test images of size \(32 \times 32\), evenly distributed across 10 classes. The training set is partitioned uniformly among all clients to form the local datasets \(\mathcal{D}_{C_i}\), and the performance of the final server model is reported on the CIFAR-10 test set. 

\subsection{Experimental Protocol}
Dataset distillation is conducted using the ConvNet architecture \cite{gidaris2018dynamic}, following the implementation design of \cite{cazenavette2022dataset}. For FL evaluation, we consider both ConvNet and AlexNet \cite{krizhevsky2012imagenet} as client and server architectures. Client model updates are aggregated via simple weight averaging; while alternative aggregation strategies may influence the quality of trajectories and the resulting synthetic data, we leave such exploration to future work.

As outlined in Algorithm~\ref{alg:alg}, the \fedsyn\ pipeline consists of three primary stages:  
(i) standard FL training for \(T\) communication rounds, during which client model trajectories are collected;  
(ii) server-side dataset distillation following the method described in Section~\ref{sec: fedsyn}, yielding the synthetic dataset \(\mathcal{D}_{\text{syn}}\); and  
(iii) adversarial training on the server for \(E^{\text{adv}}\) epochs using the synthesized data.  
Notably, the dataset distillation and adversarial training stages operate entirely on the server, imposing no additional computation or communication overhead on client devices.

\subsection{\fedsynfed~Variant}

To explore the potential benefits of incorporating \(\mathcal{D}_{\text{syn}}\) throughout the FL process, we evaluate a variant of our method in which a second full round of FL is executed after the distillation and adversarial training stages. In this setting, denoted \fedsynfed, we perform the regular FL training as in stage (i), but after each weights aggregation also perform a round of adversarial training using the \(\mathcal{D}_{\text{syn}}\) obtained from the first stage. This allows us to study the effect of repeatedly leveraging synthetic data within FL, though it necessarily doubles both client-side computation and communication costs.

\subsection{Implementation Details}
All experiments are conducted on an NVIDIA L4 GPU with 24\,GB of VRAM, a 6\,Gbps clock rate, and GDDR5 memory. We simulate an FL environment with 11 clients, where the server also participates as an additional client to supply the data required for initializing \(\mathcal{D}_{\text{syn}}\). For ConvNet, client models are trained for \(T = 60\) local epochs with an initial learning rate of 0.01, reduced to 0.001 after 40 epochs. For AlexNet, clients train for \(T = 110\) epochs with a learning rate schedule of 0.1, reduced to 0.01 after 90 epochs. All models use SGD with momentum 0.9 and weight decay \(5\times 10^{-4}\).

Dataset distillation is run for 2000 steps with a batch size of 128. Increasing the number of distilled images per class beyond 50 yields no improvement in performance, so all experiments use a distilled dataset containing 50 images per class (500 images total). This choice is also constrained by the iid.\ server data used for initialization: with 11 clients, the server holds \(\approx 5000/11\) images per class, limiting the maximum number of distilled samples per class to 454. The student model \(\mathcal{S}\) is trained for \(E = 1000\) epochs during distillation. The learning rate for updating \(\mathcal{D}_{\text{dis}}\) using Equation~\ref{eq:dist loss} is set to 1000, while \(\mathcal{S}\) is trained with a learning rate of 0.01. For the FAT baseline, we follow the implementation details provided in \cite{zizzo2020fat}.

\subsection{Evaluation and Baselines}
We report both clean accuracy and adversarial accuracy of the final global model on the CIFAR-10 test set. Adversarial robustness is evaluated using a 10-step PGD attack (\(\lambda = 10\)) with an \(\ell_p\)-norm bound of \(\epsilon = 2/255\) and step size \(\eta = 2/255\). We compare the model obtained via \fedsyn against two baselines, (i) the standard FL model trained without adversarial defenses, and (ii) FAT \cite{zizzo2020fat}, where each client performs adversarial training locally before transmitting model updates. In addition to accuracy metrics, we also measure the total wall-clock computation time incurred on each client, reflecting the practical compute burden placed on resource-constrained devices.

\begin{figure}[!t]
    \centering
    \includegraphics[scale=0.9]{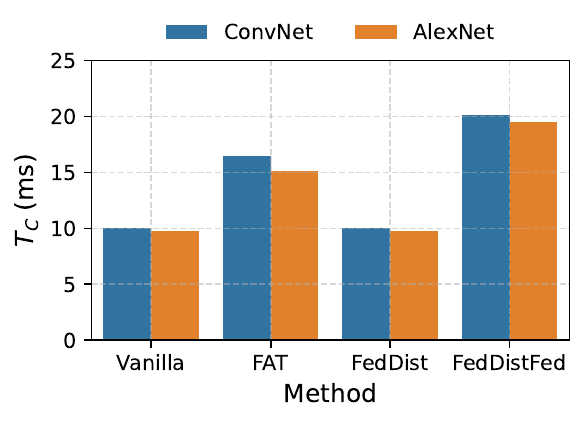}
    \caption{Comparison of time spent by one client per step in milliseconds.}
    \label{fig: plot}
\end{figure}

% \section{Results}
% \label{sec: results}
% Table~\ref{tab:results} summarizes the performance of all evaluated methods. We observe that \fedsyn improves adversarial accuracy by approximately \(10\%\) on ConvNet and \(18\%\) on AlexNet compared to the vanilla FL, while imposing \emph{no additional} computational burden on client devices. The \fedsynfed variant achieves slightly higher clean accuracy than \fedsyn, owing to the additional round of FL conducted after dataset distillation. However, this benefit comes at the cost of doubling both client-side computation and communication.

% \subsection{Discussion}
% Comparing \fedsyn with FAT shows a trade-off. While both approaches yield similar clean accuracy, FAT achieves higher adversarial accuracy but requires nearly \(60\%\) more client computation time, as also shown in Fig~\ref{fig: plot}. Importantly, the reported runtimes are measured on an NVIDIA L4 GPU; on real-world low-resource edge devices, the computational overhead of FAT would be significantly larger. As is typical in adversarial training, all defended models (\fedsyn, \fedsynfed, and FAT) show a reduction in clean accuracy relative to the vanilla model.

\begin{table}[t]
\caption{Clean and Adversarial Accuracy comparison of FedDist. $T_C$ is the time spent by one client per step in milliseconds on the NVIDIA L4 GPU}
\label{tab:results}
  \centering
  \begin{tabular}{llcccc}
    \hline
     Method& Model & $T_C$ & Clean Acc & Adv Acc & Avg\\
    \hline
    Vanilla & ConvNet & 10.07 & 74.75 & 41.73 & 58.24\\
    FAT & ConvNet & 16.45 & 70.47 & 58.96 & 64.71\\
    FedDist & ConvNet & 10.07 & 70.49 & 51.75  & 61.12\\
    FedDistFed & ConvNet & 20.14 & 72.43 & 51.71 & 62.07\\
    \hline
    Vanilla & AlexNet & 9.76 & 74.36 & 35.31 & 54.83\\
    FAT & AlexNet & 15.12 & 68.85 & 58.26 & 63.55\\
    FedDist & AlexNet & 9.76 & 66.23 & 53.09  & 59.66\\
    FedDistFed & AlexNet & 19.52 & 71.42 & 51.54 & 61.48\\  
    \hline
  \end{tabular}
%\vskip -0.1in
\end{table}

\section{Results and Discussion}
\label{results-discussions}
We evaluate the performance of all methods in terms of clean accuracy, adversarial accuracy, average accuracy, and per-step client computation time. The results are summarized in Table~\ref{tab:results}, and client runtimes are further illustrated in Fig.~\ref{fig: plot}. All runtime measurements were obtained on an NVIDIA L4 GPU, providing a consistent baseline for computational cost comparison.
Across both ConvNet and AlexNet architectures, FedDist consistently improves adversarial robustness relative to the vanilla FL baseline while maintaining the same client-side computation. On ConvNet, adversarial accuracy increases from 41.73\% to 51.75\%, and on AlexNet from 35.31\% to 53.09\%. These gains occur without increasing per-step runtime (\(T_C = 10.07~\text{ms}\) for ConvNet and \(9.76~\text{ms}\) for AlexNet), demonstrating that the distillation-based procedure enhances robustness without adding computational burden to clients. In terms of average accuracy (Avg), FedDist achieves 61.12\% on ConvNet and 59.66\% on AlexNet, indicating a more balanced performance between clean and adversarial conditions compared to the vanilla baseline (58.24\% and 54.83\%, respectively).

FedDistFed provides a modest improvement in clean accuracy over FedDist. On ConvNet, clean accuracy rises from 70.49\% to 72.43\%, and on AlexNet from 66.23\% to 71.42\%. This benefit is attributable to the additional federated optimization round conducted after dataset distillation. However, this improvement comes at the cost of doubling client-side computation and communication (\(T_C = 20.14~\text{ms}\) for ConvNet and \(19.52~\text{ms}\) for AlexNet). As shown in Fig.~\ref{fig: plot}, FedDistFed’s runtime is nearly 100\% higher than FedDist, which may limit its practical applicability in resource-constrained edge environments. The corresponding average accuracies are 62.07\% and 61.48\% for ConvNet and AlexNet, respectively, showing only marginal improvements over FedDist despite the doubled computational cost.

The comparison with FAT highlights a different trade-off between robustness and efficiency. FAT achieves the highest adversarial accuracy among all methods (58.96\% on ConvNet and 58.26\% on AlexNet) but requires substantially more computation. The per-step runtime increases by approximately 63\% on ConvNet (10.07~ms to 16.45~ms) and by 55\% on AlexNet (9.76~ms to 15.12~ms), as also depicted in Fig.~\ref{fig: plot}. FAT’s average accuracies are 64.71\% for ConvNet and 63.55\% for AlexNet, higher than both FedDist and FedDistFed. This indicates that FAT achieves a slightly better overall balance between clean and adversarial accuracy, but at a much higher computational cost. On low-resource devices, this overhead could significantly hinder practical deployment. As expected, all defended models, including FedDist, FedDistFed, and FAT, exhibit lower clean accuracy compared to the vanilla baseline.

In summary,  results indicate that FedDist provides a practical and efficient approach to improving adversarial robustness in FL, achieving a strong balance between performance and client-side computational cost.

\section{Conclusion and Future Work}
\label{conclusion-future-work}
We introduced \fedsyn, a novel framework that enables adversarial training within FL while preserving client data privacy and imposing no additional computational load on resource-constrained edge devices. Distilling a synthetic proxy of client data directly on the server, without accessing any raw local information, \fedsyn~ substantially improves adversarial robustness and highlights the broader potential of privacy-preserving dataset distillation. Beyond robustness, the ability to generate high-quality synthetic data in an FL context opens promising avenues for future research in domains, e.g., personalization, fairness, and continual learning.
The data distillation process used to generate \(\mathcal{D}_{\text{syn}}\) assumes that client datasets are drawn from an iid. distribution, an assumption that may not hold in realistic FL deployments where data heterogeneity is common. Future work may explore how to extend \fedsyn~to  non-iid. FL settings, including techniques for trajectory alignment, distribution-aware distillation, or personalized synthetic data generation.

% \section*{Acknowledgments}
% This should be a simple paragraph before the References to thank those individuals and institutions who have supported your work on this article.

\bibliographystyle{IEEEtran}
\bibliography{bare_jrnl}

\vfill
\end{document}